\documentclass[lettersize,journal]{IEEEtran}
\usepackage{amsmath,amsfonts}
\usepackage{algorithmic}
\usepackage{algorithm}
\usepackage{array}
\usepackage{textcomp}
\usepackage{stfloats}
\usepackage{url}
\usepackage{multirow} 
\usepackage{booktabs}
\usepackage{pifont}
\usepackage{verbatim}
\usepackage{graphicx}
\usepackage{cite}
\usepackage{xcolor}
\usepackage{subcaption}  
\begin{document}

\title{DACESR: Degradation-Aware Conditional Embedding for Real-World Image Super-Resolution}

\author{Xiaoyan Lei, Wenlong Zhang, Biao Luo, Hui Liang, Weifeng Cao and Qiuting Lin

\thanks{This work was supported in part by Graduate Education Reform Project of Henan Province (2023SJGLX037Y), the Research Project of Humanities and Social Sciences of the Ministry of Education, China (No. 24YJAZH075), and the Research Project of Humanities and Social Sciences of Henan Province, China (No. 2025-ZZJH-370). (Corresponding authors: Weifeng Cao; Biao Luo; Hui Liang.)}
\thanks{Xiaoyan Lei, Hui Liang and Weifeng Cao are with the  School of Electrical and Information Engineering, Zhengzhou University of Light Industry, Zhengzhou 450002, Henan, China. (e-mail: xyan\_lei@163.com; hliang@zzuli.edu.cn; weifeng\_cao@163.com)}
\thanks{Wenlong Zhang is with the Shanghai Artificial Intelligence Laboratory, Shanghai 200233, China. (e-mail: zhangwenlong@pjlab.org.cn)}
\thanks{Biao Luo is with the School of Automation, Central South University, Changsha 410083, China. (e-mail: biao.luo@hotmail.com)}
\thanks{Qiuting Lin is with the Intelligent Manufacturing Engineering at Machinery Technology Development Co.,Ltd., China Academy of Machinery Science and Technology Group Co.,Ltd., Beijing 100801, China. (e-mail: linqt2001@163.com)}

}




\maketitle

\begin{abstract}
Multimodal large models have shown excellent ability in addressing image super-resolution in real-world scenarios by leveraging language class as condition information, yet their abilities in degraded images remain limited. In this paper, we first revisit the capabilities of the Recognize Anything Model (RAM) for degraded images by calculating text similarity. We find that directly using contrastive learning to fine-tune RAM in the degraded space is difficult to achieve acceptable results. To address this issue, we employ a degradation selection strategy to propose a Real Embedding Extractor (REE), which achieves significant recognition performance gain on degraded image content through contrastive learning. Furthermore, we use a Conditional Feature Modulator (CFM) to incorporate the high-level information of REE for a powerful Mamba-based network, which can leverage effective pixel information to restore image textures and produce visually pleasing results. Extensive experiments demonstrate that the REE can effectively help image super-resolution networks balance fidelity and perceptual quality, highlighting the great potential of Mamba in real-world applications. The source code of this work will be made publicly available at: \url{https://github.com/nathan66666/DACESR.git}.
\end{abstract}


\section{Introduction}

\IEEEPARstart{S}{ingle} image super-resolution (SR) is a key challenge in computer vision and image processing, aiming to reconstruct a high-resolution image from a low-resolution input. Since the introduction of deep learning into super-resolution tasks~\cite{SRCNN}, numerous CNN-based methods have been proposed~\cite{second,FSRCNN,DRCN,VDSR,accelerating,classsr,blueprint} to enhance performance. Subsequently, the emergence of Transformer-based models~\cite{swint,swinir,HAT} has led to new breakthroughs in image super-resolution technology. These methods are based on the assumption that the degradation is known and fixed (e.g., bicubic downsampling). However, they experience a significant performance drop when the actual degradation differs from this assumption~\cite{ikc}.

Real-world Image Super-Resolution aims to restore high-resolution images under complex and unknown degradation conditions in real-world scenarios. It must handle diverse and unpredictable degradation processes, such as blurring, noise, and compression artifacts; manage data from various devices and environments; and maintain image quality robustness under suboptimal shooting conditions. Real-world image super-resolution is crucial for applications such as video surveillance image enhancement, medical imaging improvement, satellite image processing, and enhancing photos taken by smartphones.

A common approach is to synthesize real-world degradations for training real-world networks.
Significant progress has been made by introducing comprehensive degradation models to effectively synthesize realistic images~\cite{ESRGAN, bsrgan, realesrgan, swinir, dasr}. ESRGAN~\cite{ESRGAN}, which primarily uses pixel-wise loss for supervised training. BSRGAN~\cite{bsrgan} and Real-ESRGAN~\cite{realesrgan} are designed to handle severely degraded LR images. DASR~\cite{dasr} proposes a domain-gap aware training strategy and introduces a degradation space that balances a broader range of degradation levels. 
In order to restore image textures and details, most real-world image super-resolution methods rely on heavy backbone networks (e.g., RRDB~\cite{rrdb}), making them difficult to deploy on edge devices. 

Since large-scale multi-modal models possess strong priors on natural images, they have the potential to address real-world image restoration tasks. Recent methods often leverage multi-modal large models to extract visual information from low-quality images, enhancing the performance of image super-resolution networks~\cite{promptrestorer,seesr,pasd,diffbir}. Based on diffusion model backbones, these models can reconstruct natural textures and generate visually pleasing images. However, due to the inherent limitations of diffusion models, they suffer from slow inference speed and high resource consumption. Moreover, these methods lack an investigation into the recognition ability of degraded images when controlling the diffusion model.

In this paper, to enhance the ability of multi-modal large models to recognize degraded images, we first revisit the capabilities of the Recognize Anything Model (RAM)~\cite{ram} on degraded images. Specifically, we measure the model’s ability to describe image content by computing text similarity and find that as the level of image degradation increases, RAM's description accuracy significantly declines, often leading to incorrect descriptions. This issue negatively impacts the performance of super-resolution reconstruction. To address this problem, a previous study~\cite{seesr} attempted to fine-tune RAM by constructing a large degradation space. However, our analysis reveals that this approach does not effectively improve the model's ability to recognize degraded images. Therefore, in this paper, we first conduct a detailed analysis of existing methods (Section \ref{motivation}) and then propose a degradation selection strategy to train the Real Embedding Extractor (REE) using contrastive learning. This strategy enables REE to effectively extract high-level information from degraded images. Additionally, we introduce a Conditional Feature Modulator (CFM) to integrate the refined image representations from REE into the super-resolution network, thereby improving reconstruction quality. Furthermore, inspired by the recent success of Mamba in vision tasks~\cite{videossm1, videossm2, videossm3, videossm4, umamba, rsmamba, mambair, dvmsr}, we employ LAM~\cite{lam} to further evaluate its long-range modeling capabilities in real-world scenarios. Our results demonstrate that it surpasses existing non-diffusion state-of-the-art (SOTA) methods in terms of PSNR and LPIPS performance. Our contributions can be summarized as follows:

\begin{enumerate}

\item We revisit the capabilities of the RAM on degraded images and propose a method to measure its accuracy in image content description.

\item We propose the Real Embedding Extractor (REE), which improves the recognition accuracy of the RAM on all degradation types by leveraging a degradation selection strategy.

\item We extend the Mamba super-resolution network to real-world image super-resolution, uncovering the potential of Mamba-based networks in real-world image super-resolution.

\item We incorporate high-level information of REE into a powerful Mamba-based network through a Conditional Feature Modulator (CFM), achieving state-of-the-art performance in real-world image super-resolution.

\end{enumerate}

\section{Related Work}
\subsection{Non-blind Image Super-Resolution}
In non-blind image super-resolution, the low-resolution image is generated from a high-resolution image using a known degradation process. Deep learning was first applied to image super-resolution by~\cite{SRCNN}, which improves performance by learning the mapping from low-resolution to high-resolution images.~\cite{VDSR} introduces a deeper network structure, enhancing image restoration accuracy through residual learning while also accelerating convergence. Models like~\cite{swinir,HAT,hipa}, which use Transformer-based architectures, have demonstrated the strong performance of Transformers in low-level tasks. Methods such as those in~\cite{srgan,ESRGAN,zhang2019ranksrgan,zhang2021ranksrgan} use generative adversarial networks (GANs) to enhance the model's ability to restore image details and improve image quality. FDSR~\cite{fdsr} proposes an SR network based on frequency division, which processes images in the frequency domain to make the reconstruction process more transparent and interpretable. In recent years, designing lightweight image super-resolution models has become an important focus. Methods like~\cite{FSRCNN,VDSR,LapSRN,DRRN,RFDN} are representative CNN-based efficient SR approaches, striking a balance between efficiency and effectiveness. 
CFSR~\cite{cfsr} uses large-kernel convolution as a feature mixer to replace the self-attention module, inheriting the advantages of both convolution-based and transformer-based methods. Additionally,~\cite{dvmsr} has demonstrated the potential of Mamba-based networks in lightweight image super-resolution.

\subsection{Real-World Image Super-Resolution}

Real-world Image Super-Resolution aims to restore high-resolution images under complex and unknown degradation conditions in real-world scenarios. 
Several GAN-based methods have been proposed for super-resolution, among which SRGAN~\cite{srgan} is widely regarded as a seminal work. It introduced adversarial training into the super-resolution task and was capable of reconstructing high-resolution images with realistic texture from $4\times$ downsampled inputs. ESRGAN~\cite{ESRGAN} introduces Residual-in-Residual Dense Blocks (RRDB) without batch normalization as the fundamental network building unit, achieving superior visual quality. RankSRGAN~\cite{zhang2019ranksrgan, zhang2021ranksrgan} proposes a ranking-based super-resolution generative adversarial network, which optimizes the generator using perceptual metric-guided ranking. 
In recent years, diffusion models have begun to attract attention in image restoration tasks. For example, StableSR~\cite{stablesr} designs a time-aware encoder to control the Stable Diffusion process. PASD~\cite{pasd} introduces a PACA module that effectively injects pixel-level conditional information into the diffusion prior, thereby achieving higher fidelity. DiffBIR~\cite{diffbir} formulates the image restoration problem as a two-stage process, leveraging the generative capability of latent diffusion models in the second stage to produce realistic details. Despite their remarkable success, these methods suffer from high training costs, slow inference speed, and difficulties in lightweight deployment. Furthermore, in real-world scenarios, researchers have also been exploring areas such as degradation modeling~\cite{closerlookBSR,realesrgan,bsrgan}, multi-task learning~\cite{tgsr}, and systematic evaluation~\cite{zhang2023seal}.
For degradation modeling, classic blind SR methods~\cite{zhang2018learning} primarily use Gaussian blur and noise to simulate the distribution of real images. BSRGAN~\cite{bsrgan} introduces a comprehensive degradation model that incorporates multiple degradations using a shuffling strategy, while RealESRGAN~\cite{realesrgan} adoptes a higher-order strategy to construct a large degradation model. DASR~\cite{dasr} employes a three-level degradation distribution (i.e., two first-order and one higher-order degradation model) to simulate the distribution of real images. Although random combinations can enhance robustness, the resulting degradation chain does not correspond to a real physical process and fails to cover all possible degradations, leading to a domain gap during model training.


\subsection{Conditional Network}
Real-world image super-resolution is challenging due to unpredictable noise and degradation. A conditional network can guide the model to focus on different aspects of an image (e.g., texture, sharpness, or noise reduction) depending on the condition provided. This helps in producing more realistic and high-quality super-resolution results. Many previous works have achieved significant success in image super-resolution tasks by introducing conditional networks. For example,~\cite{dasr} incorporates conditional inputs to adjust the super-resolution process based on different degradation domains.~\cite{adafm} introduces channel-wise feature modification, allowing a model to accurately adapt to varying restoration levels. By leveraging conditional networks, AdaFM effectively modulates model parameters to handle different levels of image degradation. 
Similarly, \cite{csenet} introduces Global Feature Modulation, which enhances model flexibility by using conditional networks to extract global features for image retouching. Additionally, \cite{stablesr} employs conditional GANs with multiple latent codes to handle diverse degradation scenarios, enabling the model to restore high-quality images. \cite{pasd} introduces a pixel-aware cross-attention module, enabling the diffusion model to perceive pixel-level local image structures. These studies show that conditional networks improve the flexibility and performance of super-resolution models, especially in complex real-world degradation situations. Improving the accuracy of conditional information and fully leveraging such information remain critical challenges that demand further attention, and also constitute the primary motivation of this paper.

\begin{figure*}[!htbp]
\centering
\includegraphics[width=\textwidth]{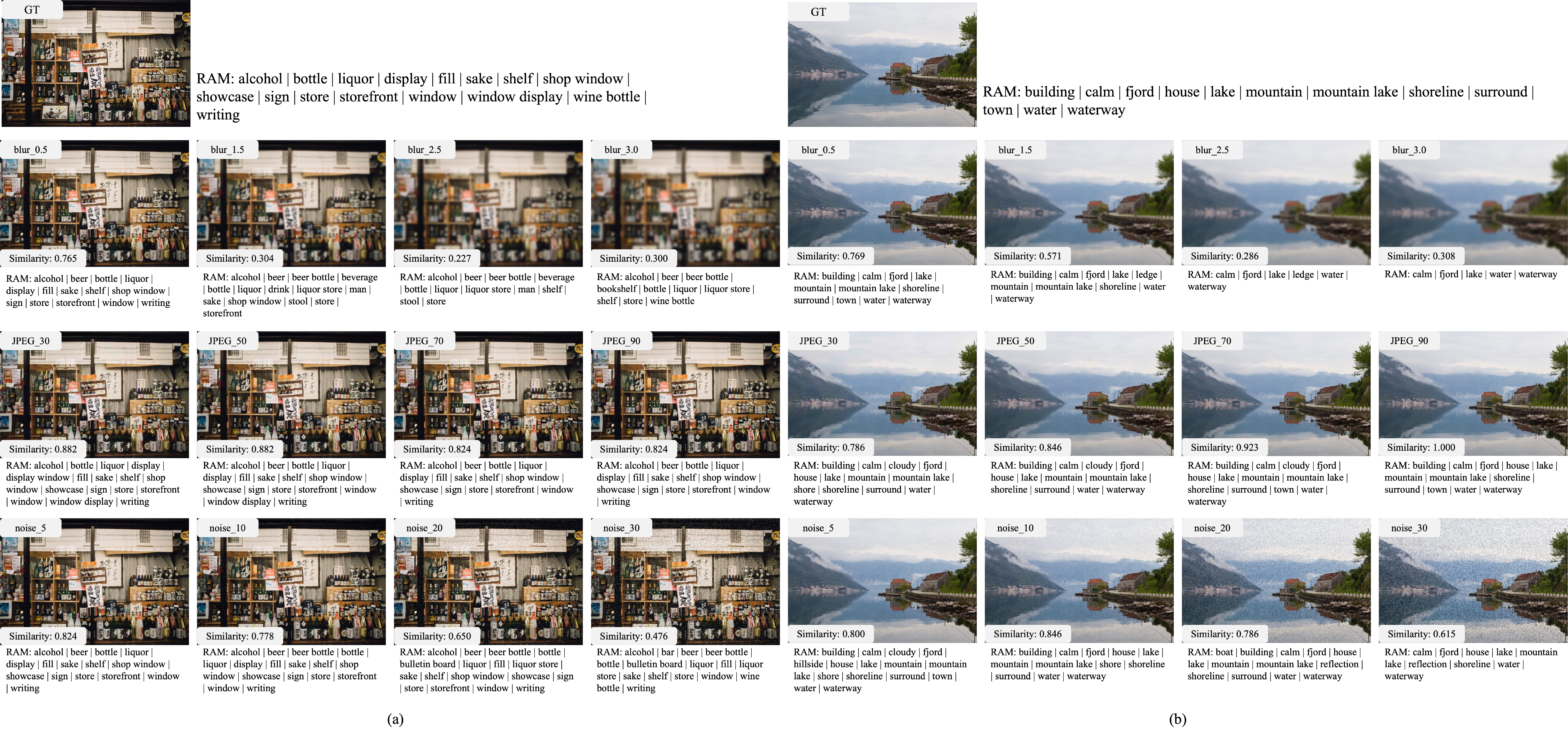}
\caption{The tag representations of RAM on clean images and images with varying levels of degradation.``Similarity" refers to the Jaccard similarity (Eq.~(\ref{eq:sim})).}
\label{fig:ram-all}
\end{figure*}
\section{Motivation}\label{motivation}

\begin{figure*}[ht]
    \centering
    \begin{minipage}{\textwidth}
        \centering
        \begin{subfigure}{0.35\textwidth}
            \centering
            \includegraphics[width=\linewidth]{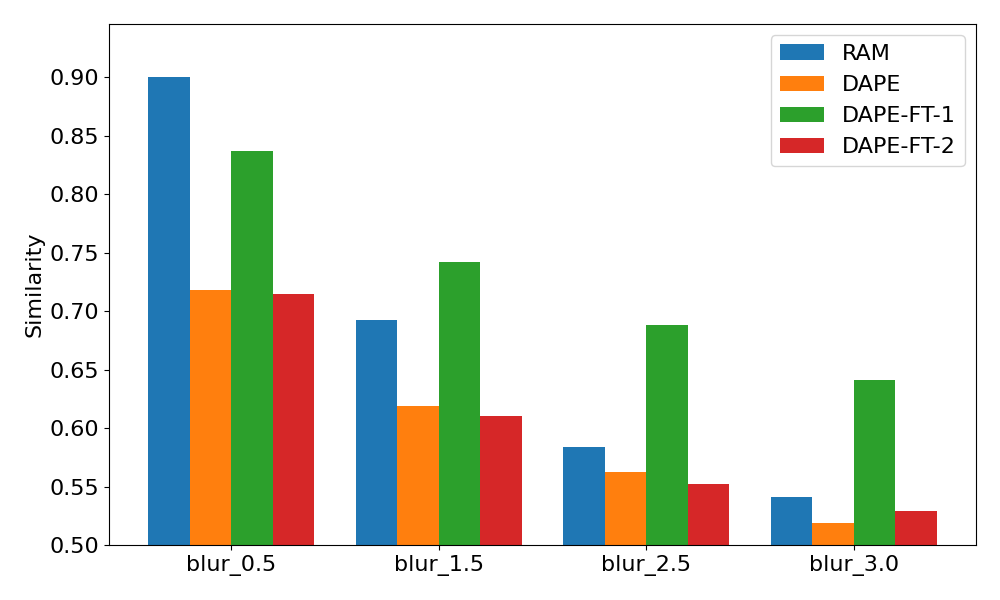}  
            \caption{Blur degradation}
            \label{fig:subfig1}
        \end{subfigure}
        \hspace{2cm} 
        \begin{subfigure}{0.35\textwidth}
            \centering
            \includegraphics[width=\linewidth]{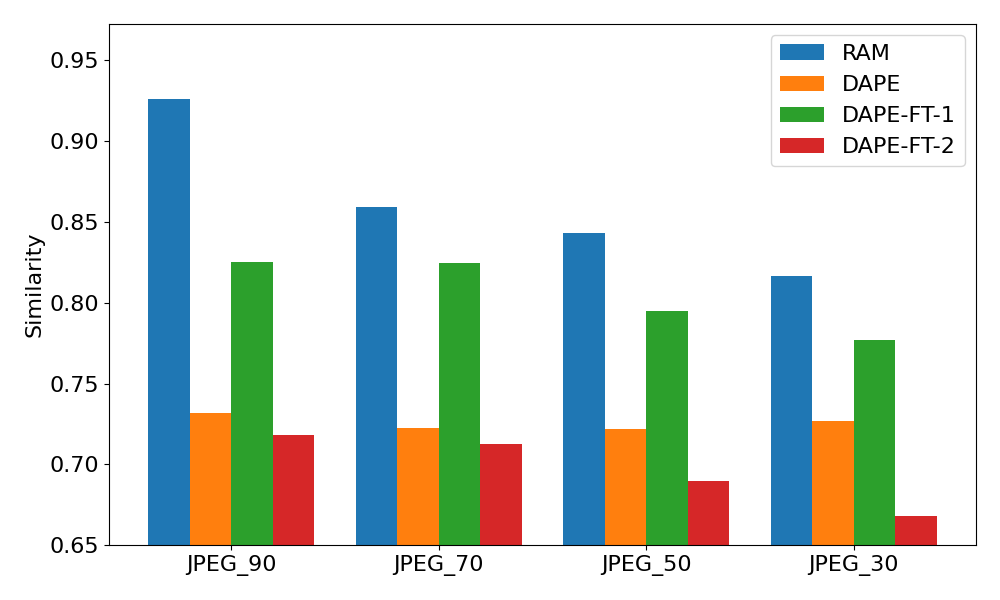} 
            \caption{JPEG degradation}
            \label{fig:subfig2}
        \end{subfigure}
    \end{minipage}
    \\[1em]
    \begin{minipage}{\textwidth}
        \centering
        \begin{subfigure}{0.35\textwidth}
            \centering
            \includegraphics[width=\linewidth]{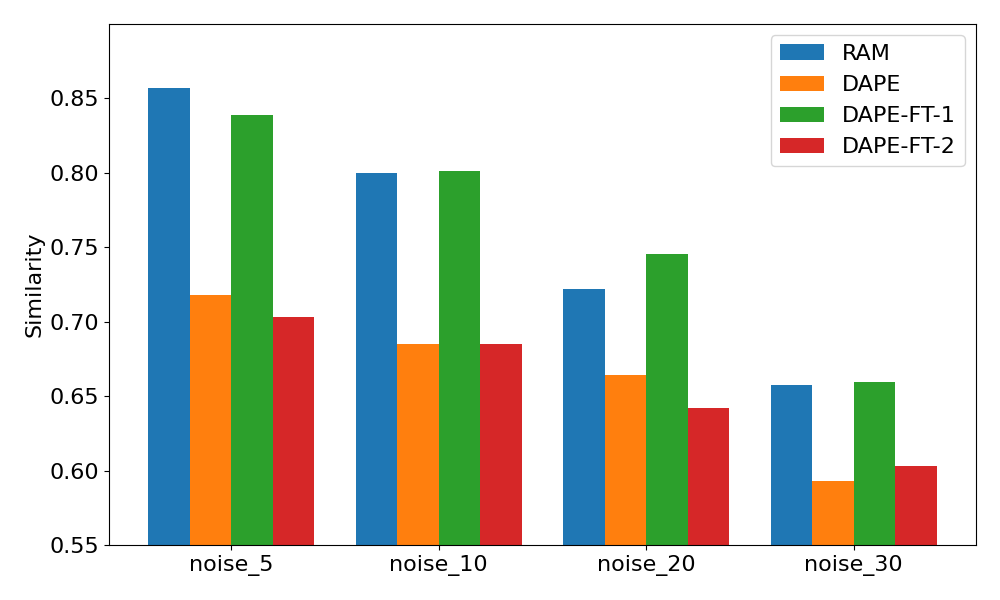} 
            \caption{Noise degradation}
            \label{fig:subfig3} 
        \end{subfigure}
        \hspace{2cm} 
        \begin{subfigure}{0.35\textwidth}
            \centering
            \includegraphics[width=\linewidth]{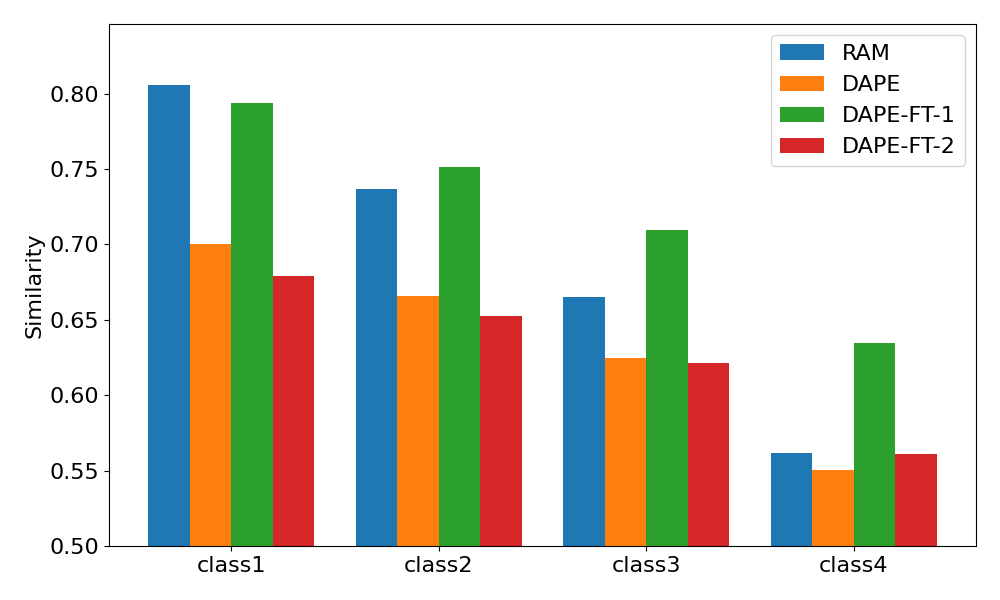} 
            \caption{Multiple degradation}
            \label{fig:subfig4} 
        \end{subfigure}
    \end{minipage}
    \caption{Comparison of text output accuracy across RAM under different types and intensities of degradation. (a) Blur: The x-axis represents isotropic Gaussian blur, where larger values indicate stronger blurring. (b) JPEG: The x-axis denotes JPEG compression levels, with lower values indicating higher compression. (c) Noise: The x-axis represents the intensity of additive Gaussian noise, where higher values correspond to increased noise levels. In (d), the classification is based on the text similarity values of RAM for different degraded outputs, which are evenly divided into four categories in descending order. Each category contains multiple types/levels of degradation.}
    \label{fig:cls}
\end{figure*}

\subsection{Re-exploring the Capabilities of Recognize Anything Model on Degraded Images}

To explore the potential of super-resolution models in real-world scenarios, we use pre-trained multimodal models to encode degraded images and help the super-resolution network recover more realistic details. We expect the Recognize Anything Model (RAM), with its rich prior knowledge, to mitigate the effects of degradation by providing clean encodings or accurate content descriptions. As shown in Fig.~\ref{fig:ram-all}, RAM tags the content of images at different degradation levels. While it accurately describes ground truth images, its descriptive ability varies for degraded images. We examine two sets of images, as shown in Fig.~\ref{fig:ram-all}(a)(b).
We observe that RAM's descriptive accuracy seems to be related to the level of image degradation. To quantify the relationship between pre-trained models and image degradation, similarity metrics (i.e., Jaccard similarity) can be used to evaluate how well the model's output (i.e., tag descriptions) aligns with the ground truth or clean image under different levels of degradation. The similar calculation method can be expressed as:
\begin{equation} \label{eq:sim}
  \begin{split}
  J(A, A') = \frac{|A \cap A'|}{|A \cup A'|}
  \end{split}
\end{equation}
where \( A \) represents the description of a clean image, and \( A' \) represents the description of an image with degradation but with the same content as \( A \). \( |A \cap A'| \) represents the number of common elements in the content descriptions of images \( A \) and \( A' \), while \( |A \cup A'| \) represents the total number of unique elements in the content descriptions of images \( A \) and \( A' \). The larger the value of \( J(A, A') \), the higher the similarity between the two images, and vice versa.
In Fig.~\ref{fig:ram-all}, it can be observed that images with more severe visual degradation tend to have lower ``Similarity" values. For example, in Fig.~\ref{fig:ram-all}(a), as the noise level increases (e.g., noise\_5, noise\_10, noise\_20, noise\_30), the similarity gradually decreases. However, not all types of degradation follow this trend. For instance, in (a), JPEG compression does not show a clear trend across the four levels, and visual inspection shows no significant degradation, indicating that higher similarity reflects closer resemblance to the ground truth. In contrast, for blur degradation, as the blur level increases, text similarity decreases. Although the similarity score for blur\_2.5 is slightly lower than for blur\_3.0, both represent severe degradations with low text similarity scores.
In Fig.~\ref{fig:ram-all}(b), we observe that the same degradation results in different text similarity values depending on the image content, but it does not affect the overall trend. 
In Section~\ref{ob2}, we propose a quantitative evaluation method that mitigates the interference of image content by averaging while emphasizing the impact of degradation on the ability of RAM to recognize image content. Through the charts in Figs.~\ref{fig:cls}(a),~\ref{fig:cls}(b),~\ref{fig:cls}(c) and~\ref{fig:cls}(d), we observe that as the degradation intensity increases, the descriptive ability of RAM for images declines.
Therefore, Higher textual similarity indicates better image quality, while lower similarity reflects poorer quality, especially as degradation intensity increases.

\subsection{Enhancing the Representation Capability of Recognize Anything Model for Degraded Images}
\label{ob2}

Due to the limited descriptive capability of RAM for degraded images, \cite{seesr} introduces DAPE, which is fine-tuned from a pre-trained tag model, i.e., RAM. It forces the representation embedding and logits embedding from the LR branch to be close to those of the HR branch, making DAPE robust to image degradation.

\noindent\textbf{A quantitative method for evaluating the severity of degradation.} To validate the robustness of this approach against image degradation, firstly, we select $n$ HR images from the DIV2K dataset and use the degradation model from~\cite{realesrgan} to generate $m$ types of degraded LR images. Secondly, we apply RAM and DAPE to these images to generate corresponding text descriptions $T$, and calculate the similarity between the tags of each degraded image and the ground truth. Since similarity can be influenced by image content, we took the average similarity $S$, across images with the same type of degradation. This process can be expressed as follows:
\begin{equation} \label{eq:dape}
  \begin{split}
    & T_{r}^{i,j} = RAM(I_{LR}^{i,j}),i\in 0,1,\cdots ,m, j\in 0,1,\cdots ,n, \\
    & T_{d}^{i,j} =DAPE(I_{LR}^{i,j}),i\in 0,1,\cdots ,m, j\in 0,1,\cdots ,n,  \\
    & T_{g}^{j} = RAM(I_{GT}^{j}),j\in 0,1,\cdots ,n, \\
    & S_{r}^{i} = \frac{1}{n} {\textstyle \sum_{j=0}^{n}} J(T_{g}^{j}, T_{r}^{i,j}),i\in 0,1,\cdots ,m, j\in 0,1,\cdots ,n,  \\
    & S_{d}^{i} = \frac{1}{n} {\textstyle \sum_{j=0}^{n}} J(T_{g}^{j}, T_{d}^{i,j}),i\in 0,1,\cdots ,m, j\in 0,1,\cdots ,n.
  \end{split}
\end{equation}
where \( T_{r}^{i,j} \) represents the tag generated by RAM for the \( j \)-th LR image with the \( i \)-th type of degradation, and \( T_{d}^{i,j} \) represents the tag generated by DAPE for the \( i \)-th LR image with the \( j \)-th type of degradation. \( T_{g} \) denotes the text description of the ground truth image generated by RAM. \( S_{r}^{i} \) denotes the average similarity of text descriptions generated by RAM for the \( i \)-th type of degradation across $n$ images, while \( S_{d}^{i} \) represents the average similarity of text descriptions generated by DAPE for the \( i \)-th type of degradation across $n$ images. The calculation method for \( J(\cdot, \cdot) \) can be found in Eq. (\ref{eq:sim}).

\noindent\textbf{DAPE performs poorly on images with various types of degradation.} We sort the randomly generated degradations based on the text similarity values output by RAM, categorizing the degradations into four levels from mild to severe(class1, class2, class3 and class4). As shown in Fig.~\ref{fig:cls}(d), the text similarity output by RAM decreases as the degradation intensifies. 
Similarly, DAPE's text similarity decreases across four degradation levels, but it consistently remains lower than RAM at each level, suggesting DAPE has not effectively improved RAM's ability to recognize degraded image content.
To better understand this relationship, we explore three types of degradation with varying severity levels. As shown in Figs.~\ref{fig:cls}(a),~\ref{fig:cls}(b), and~\ref{fig:cls}(c), the text output accuracy of DAPE is lower than that of RAM across different types and levels of degradation.

\begin{figure*}[!htbp]
\centering
\includegraphics[width=0.9\textwidth]{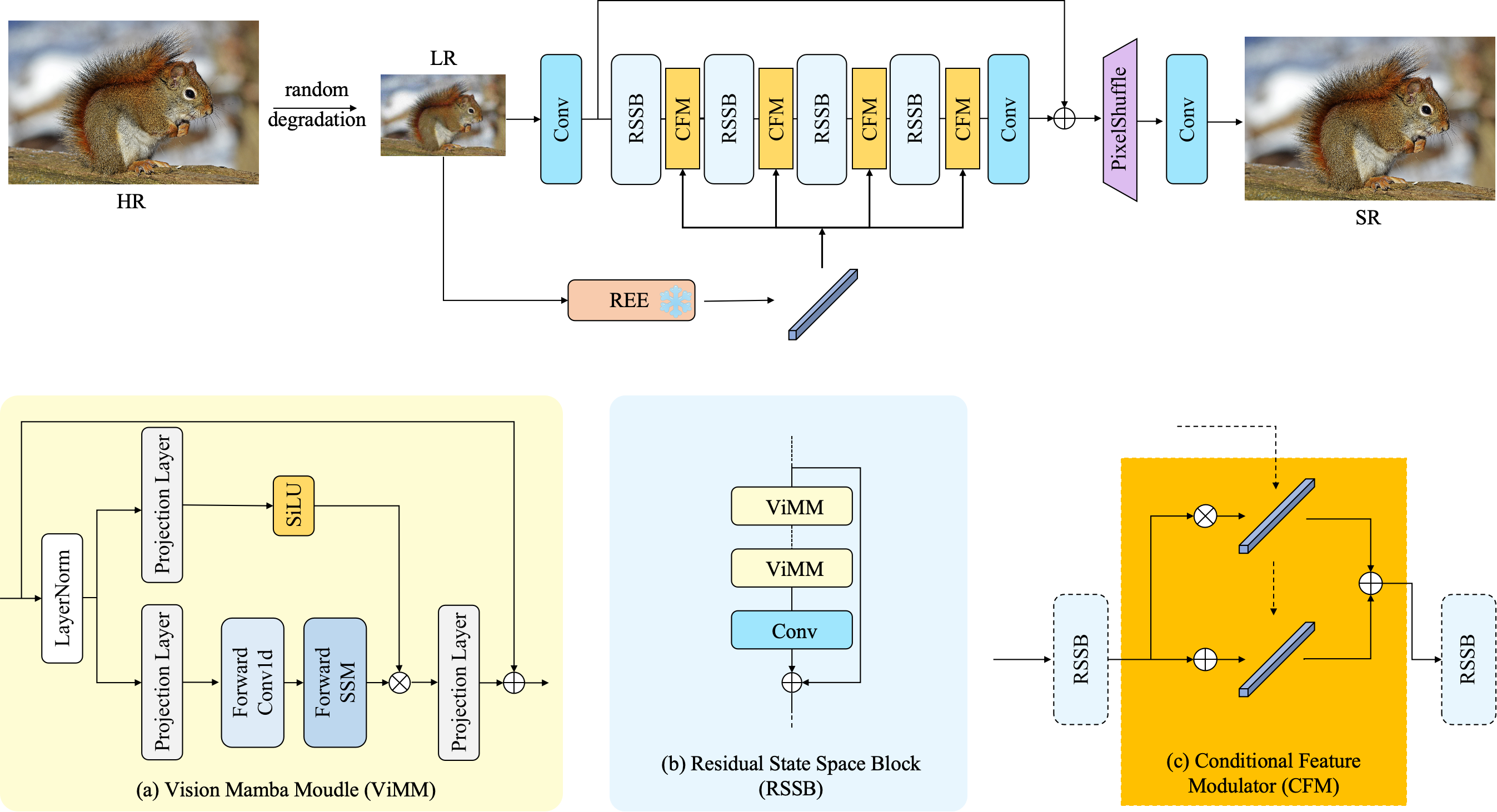}
\caption{The overview of DACESR.}
\label{fig:dacnsr1}
\end{figure*}

\begin{figure*}[!htbp]
\centering
\includegraphics[width=0.9\textwidth]{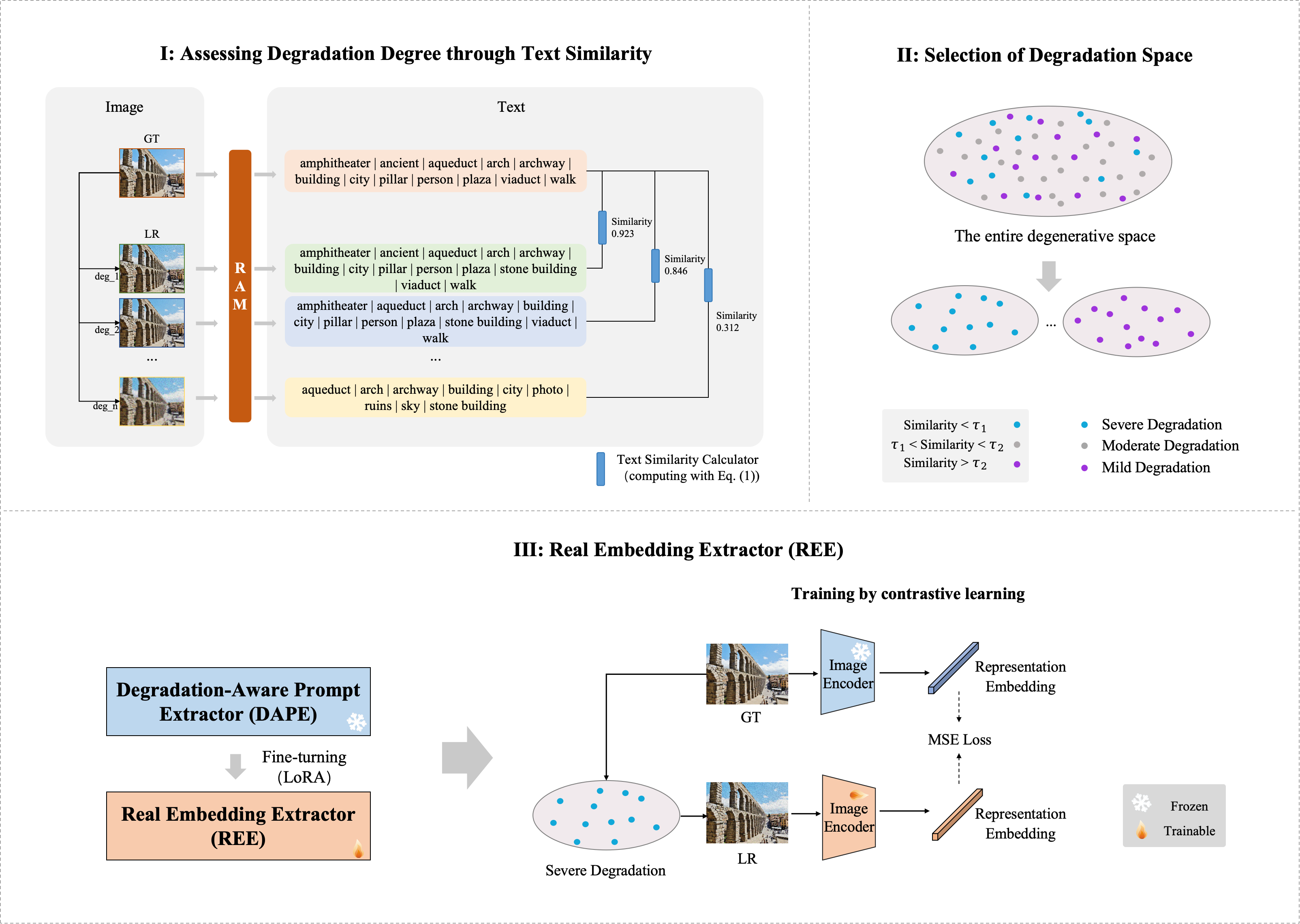}

\caption{The training pipeline of the Real Embedding Extractor (REE).}
\label{fig:dacnsr2}
\end{figure*}



\noindent\textbf{The grouped training strategy can effectively improve DAPE performance.} We seek to explore the underlying reasons for this phenomenon. Inspired by~\cite{tgsr}, we attempt to mitigate the issue by employing group-based training.

We fine-tune DAPE using all mild degradations from class1 and all severe degradations from class4. Here, ``DAPE-FT-1" represents DAPE fine-tuned with severe degradations, while ``DAPE-FT-2" represents DAPE fine-tuned with mild degradations. As shown in Fig.~\ref{fig:cls}(d), DAPE-FT-1 achieves higher image description accuracy across all degradation categories compared to DAPE. For mild degradations (class1 and class2), DAPE-FT-1 performs comparably to RAM, while for severe degradations (class3 and class4), it outperforms both RAM and DAPE in textual similarity. Conversely, DAPE-FT-2 shows suboptimal performance across all degradation levels, regardless of severity. 
In Figs.~\ref{fig:cls}(a),~\ref{fig:cls}(b),~\ref{fig:cls}(c), DAPE-FT-1 shows greater robustness to varying levels and types of degradation compared to DAPE. It performs slightly lower than RAM on JPEG, blur\_0.5, and noise\_5, which are mild degradations with good visual quality.
Meanwhile, DAPE-FT-2 performs similarly to DAPE on Blur and Noise but fell behind on JPEG and underperforms RAM across all categories and degradation levels. Thus, training exclusively on severe degradations appears to improve the model's overall robustness, while training on mild or mixed degradations weakens performance. This may be due to: 1) Training on heavily degraded images makes the model more inclined to ignore noise, blur, and other artifacts, allowing it to focus on global and local key features (such as object structures and semantic information). As a result, it can still recognize image content when faced with mild degradation. 2) For mixed degradation, due to the lack of explicit distribution modeling, the model may fail to learn appropriate transformation rules between different levels of degradation, making it difficult to adapt to various degrees of degradation. 3) Mildly degraded images are usually closer to high-quality real images compared to heavily degraded images. As a result, the model may overfit these relatively ``clean" samples and lack the ability to handle more severe degradation.

\section{Methodology}
\subsection{Preliminaries}
State space models (SSMs), such as the Mamba deep learning model, hold potential for long sequence modeling. Inspired by continuous systems, SSMs map a 1-D function or sequence $x(t)\in \mathbb{R} \longmapsto y(t)\in \mathbb{R}$ via a hidden state $h(t)\in \mathbb{R}^N $. The formulation is as follows:
\begin{equation} \label{eq1}
  \begin{split}
  & h'(t)=Ah(t)+Bx(t), \\
  & y(t)=Ch(t).
  \end{split}
\end{equation}
where N is the state size, $A\in \mathbb{R}^{N\times N} $, $B\in \mathbb{R}^{N\times 1} $, $C\in \mathbb{R}^{1\times N} $. 

Mamba is the discrete version of the continuous system, and it achieves this by utilizing $\Delta$  to convert continuous parameters A and B into their discrete counterparts, $\bar{A}$ and $\bar{B}$.  The commonly used method for transformation is zero-order hold (ZOH), which is defined as follows:
\begin{equation} \label{eq2}
  \begin{split}
  & \bar{A} =exp(\Delta A),\\
  & \bar{B}=(\Delta A)^{-1}(exp(\Delta A)-I)\cdot \Delta B.
  \end{split}
\end{equation}

After the discretization of $\bar{A}$, $\bar{B}$, the discretized version of Eq.~\ref{eq1} using a step size $\Delta$ can be rewritten as:
\begin{equation} \label{eq4}
  \begin{split}
  & h_t=\bar{A}h_{t-1}+\bar{B}x_t, \\
  & y_t=Ch_t.
  \end{split}
\end{equation}

\subsection{Overall of DACESR}

The network architecture consists of two main parts: a super-resolution network and a Real Embedding Extractor (Fig.~\ref{fig:dacnsr1}). The low-resolution image is processed through two parallel branches. In the main branch, the image is passed through convolutional layers and Residual State Space Blocks (RSSBs)~\cite{dvmsr} to extract deep features. 
In the conditional branch, the low-resolution image is fed into the Real Embedding Extractor (REE) to obtain the correct image representation, which is then integrated into each deep feature layer of the super-resolution network through Conditional Feature Modulator (CFM).
The final features are summed up to produce the input for the high-quality reconstruction stage, which generates the super-resolution output image.

\subsubsection{Real Embedding Extractor} 

The Real Embedding Extractor (REE) is designed to extract the correct image representation, as illustrated in Fig.~\ref{fig:dacnsr2}. Based on our previous analysis, we observed that as image degradation intensifies, the Recognize Anything Model (RAM)'s ability to accurately describe image content deteriorates. To address this, we utilize the degradation space proposed in~\cite{realesrgan} and select a subset of images from the DIV2K training set. Using the calculation method in Eq.~(\ref{eq:dape}), we measure the text similarity of RAM’s output descriptions and sort them in descending order. The thresholds $\tau_1$ and $\tau_2$ are set, where images with similarity scores above $ \tau_1 $ are categorized as mildly degraded, while those below $ \tau_2 $ are considered severely degraded. In our approach, COCO~\cite{coco} is used as the training dataset, and severe degradation is randomly added during training. Additionally, we fine-tune DAPE~\cite{seesr} using LoRA~\cite{lora}. This process is illustrated in Fig.~\ref{fig:dacnsr2}.  Through contrastive learning, REE can correct the erroneous representations of degraded images. When a low-resolution image passes through REE, it produces an encoded representation that closely approximates that of a high-resolution image. This refined representation assists the super-resolution network in reconstructing images in real-world scenarios more accurately.

\subsubsection{Conditional Feature Modulator} 
The overall framework is depicted in Fig.~\ref{fig:dacnsr1}. The Conditional Feature Modulator (CFM) adaptively adjusts input features by incorporating conditional information, where the input features are split into two streams that undergo multiplication and addition operations with the conditional information to control the scale and shift of the features. In this way, the CFM module can dynamically adjust the output features based on different conditions, enhancing the model's adaptability and performance in tasks such as multimodal learning or conditional generation. 
CFM can be formulated as: 
\begin{equation} \label{eq3}
  \begin{split}
  & CFM(x_{i}) = \alpha \cdot  x_{i} + \beta, \\
  & \alpha = R\&I(x_{c}),\\
  & \beta = R\&I(x_{c'}),\\
  \end{split}
\end{equation}
where $x_{i}$ represents the feature map output of the i-th RSSB. $R\&I$ represents the resize and interpolation operation, and $x_{c}$ and $x_{c'}$ are the feature representations of the conditional information. In this paper, \( c \) and \( c' \) are equal.

\begin{table*}[!htbp]
    \centering
    \caption{Quantitative comparison with state-of-the-art methods for real-world image super-resolution on benchmark datasets. For the compared methods, we employ their officially released pre-trained models. The best and second best results are highlighted in {\color{red}red} and {\color{blue}blue} respectively.}
    \resizebox{\linewidth}{!}{
    \begin{tabular}{llcccccccccc}
    \hline
    \multirow{2}{*}{Datasets} & \multirow{2}{*}{Metric} & ESRGAN  & BSRGAN  & Real-ESRGAN  & SwinIR  & DASR  & HAT & {KDSR} & {DCLS} & {CDFormer} & DACESR  \\
     & &  \cite{ESRGAN} & \cite{bsrgan} &  \cite{realesrgan} &  \cite{swinir} &  \cite{dasr} & \cite{HAT}  & {\cite{kdsr}} & {\cite{dcls}} &  {\cite{cdformer}} & (Ours) \\
    \hline
    \hline
    \multirow{2}{*}{Bicubic} & $\mathrm{PSNR}~(\uparrow)$  & 28.17 & 27.32 & 26.65 & 27.21 &{28.55} & 27.44& 27.48& {\color{blue}30.73} &  {\color{red}30.87} & {28.34}\\
                             & $\mathrm{LPIPS}~(\downarrow)$ & {\color{red}0.115} &0.236 &0.228 & 0.214& 0.170& 0.203 & 0.295 & 0.291 &  0.286 &{\color{blue}0.139}\\\hline
    \multirow{2}{*}{Level-1} & $\mathrm{PSNR}~(\uparrow)$  & 21.16 & 26.78 & 26.17 &26.45 & {\color{blue}27.84} & 26.85 & 26.69 & 27.32 & 26.90 &{\color{red}28.16} \\
                             & $\mathrm{LPIPS}~(\downarrow)$ & 0.473 &0.241 &0.231 & 0.216& {\color{blue}0.171} &0.202 & 0.307 & 0.368 &  0.363 & {\color{red}0.142}\\\hline
    \multirow{2}{*}{Level-2} & $\mathrm{PSNR}~(\uparrow)$  & 22.77 &26.75 & 26.16& 26.39&{\color{blue}27.58} &26.81 & 26.76 & 27.24 &  26.92 & {\color{red}27.68}\\
                             & $\mathrm{LPIPS}~(\downarrow)$ & 0.490 & 0.246& 0.239& 0.221&0.213 &{\color{blue}0.210}  & 0.318 & 0.416 &  0.415& {\color{red}0.171}\\\hline
    \multirow{2}{*}{Level-3} & $\mathrm{PSNR}~(\uparrow)$  & 23.63 &{24.05} &23.81 & 23.46&{23.93} & {\color{red}24.20}& {\color{blue}24.10} & 23.35 & 23.64 &23.70 \\
                             & $\mathrm{LPIPS}~(\downarrow)$ & 0.731 &0.400 &0.390 &0.377 &0.414 & {\color{red}0.331}& 0.425 & 0.627 &  0.587& {\color{blue}0.356}\\\hline
    \multirow{2}{*}{RealSR-cano~\cite{real-SR-cano}} & $\mathrm{PSNR}~(\uparrow)$ & {27.67} & 26.91& 26.14 & 26.64 & {27.40} & 26.68& 27.16 & {\color{red}27.87} &  {\color{blue}27.85} & 27.25 \\
                                  & $\mathrm{LPIPS}~(\downarrow)$ & 0.412 & 0.371&0.378 &0.357 &0.393 &{\color{blue}0.342} & 0.375 & 0.404 &  0.404&{\color{red}0.283} \\\hline
    \multirow{2}{*}{RealSR-Nikon~\cite{real-SR-cano}} & $\mathrm{PSNR}~(\uparrow)$ & {27.46} & 25.56&25.49 & 25.76& {26.35}& 25.85& 26.65 & {\color{red}27.50} &  {\color{blue}27.49}& 26.06\\
                                   & $\mathrm{LPIPS}~(\downarrow)$ & 0.425 & 0.391& 0.388& 0.364 &0.401 &{\color{blue}0.358}  & 0.383 & 0.416 &  0.417& {\color{red}0.288}\\\hline
    \multirow{2}{*}{AIM2019-val~\cite{AIM2019-val}}  & $\mathrm{PSNR}~(\uparrow)$ & 23.16 & {\color{blue}24.20}&23.89 &23.89 & 23.76& {24.19} & 24.21 & {\color{red}24.26} &  {\color{blue}24.20} &24.01 \\
                                   & $\mathrm{LPIPS}~(\downarrow)$ & 0.550 &0.400 &0.396 & 0.387&0.421 &{\color{blue}0.370}  & 0.375 & 0.535 &  0.538&{\color{red}0.252}\\\hline
    \hline
    \end{tabular}}
    \label{tab:real}
\end{table*}

\subsubsection{Mamba Network} The design of the Mamba network is shown in Fig.~\ref{fig:dacnsr1} (a). The Vision Mamba Module (ViMM) using unidirectional sequence modeling. 
The input token sequence $X \in \mathbb{R}^{H \times W \times C}$ is first normalized by the normalization layer. The normalized sequence is then linearly projected, with its feature channels expanded to $\lambda C$. Subsequently, the projection layer is processed through a 1-D convolution to compute $X_1$ via the SSM. Finally, $X_1$ is gated by the projection layer and combined with a residual connection to obtain the output token sequence $X_{\text{out}} \in \mathbb{R}^{H \times W \times C}$.


\begin{figure}[!htbp]
\centering
\includegraphics[width=0.48\textwidth]{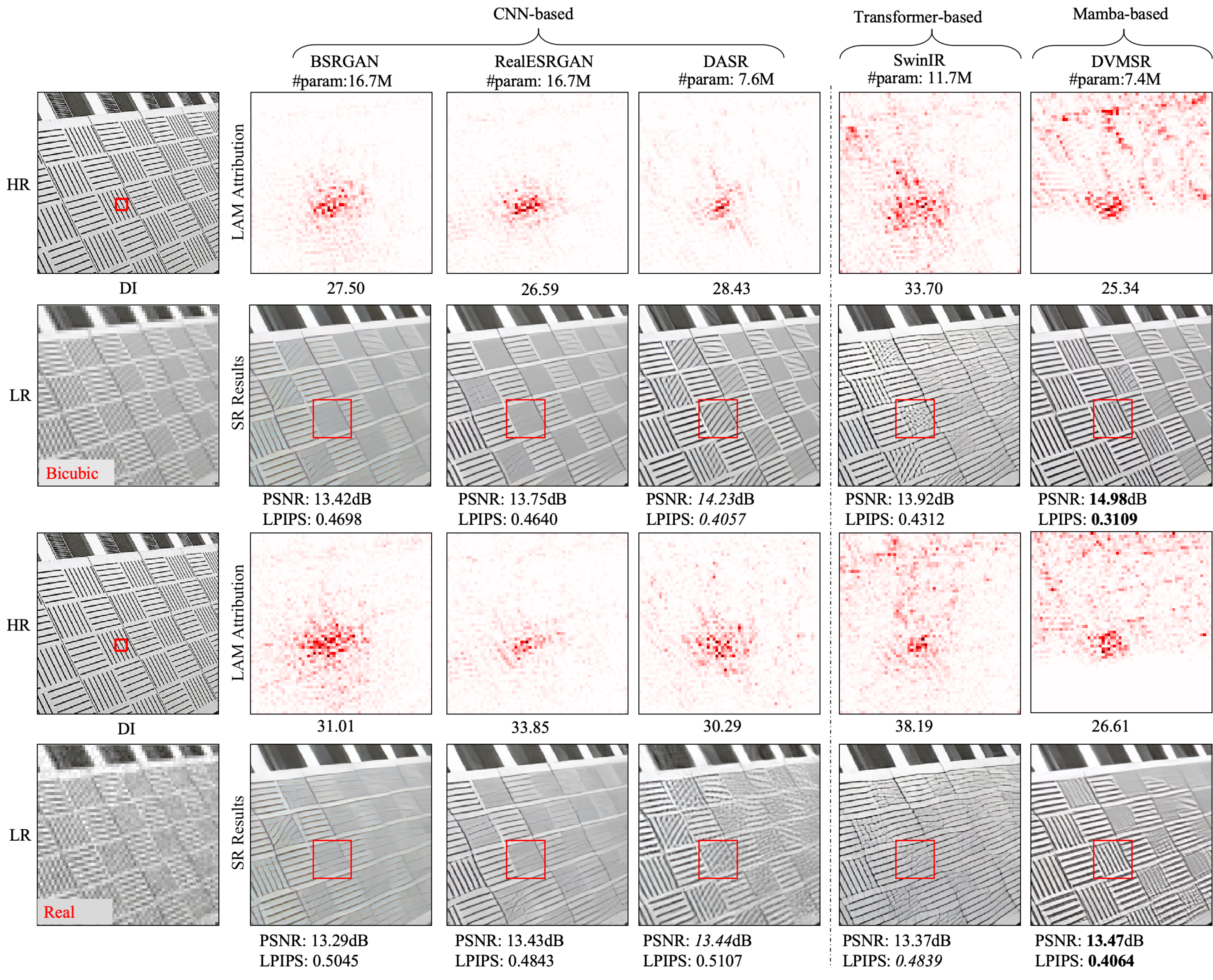}
\caption{The LAM results of different model architectures across various types of degradation. LAM attribution indicates the significance of each pixel in the input LR image during the reconstruction process of the patch highlighted by a box. The Diffusion Index (DI) denotes the extent of pixel involvement. A higher DI indicates a broader range of utilized pixels.}
\label{fig:lam}
\end{figure}
\noindent\textbf{Motivation of Mamba Network}. Motivated by Mamba’s long-range modeling capabilities~\cite{dvmsr}, we investigate its performance in real-world 
 image super-resolution tasks, comparing it to CNN-based SR methods~\cite{bsrgan, realesrgan, dasr} and transformer-based method~\cite{swinir}. To elucidate Mamba's operational mechanisms, we employed a specialized diagnostic tool called LAM~\cite{lam}, designed specifically for SR tasks. Utilizing LAM enabled us to pinpoint the input pixels that contribute most significantly to the selected region. As depicted in Fig.~\ref{fig:lam}, the red-marked points denote informative pixels crucial for the reconstruction process. 
By comparing LAM results across different degradations and architectures, we find that SwinIR~\cite{swinir} uses more pixels for reconstruction but often produces cluttered textures. In contrast, DVMSR~\cite{dvmsr} consistently uses the fewest pixels across various degradations but achieves clear textures and good visual quality. This suggests that using more pixel information does not necessarily lead to better reconstruction. DVMSR does not focus on the quantity of pixels but instead selectively attends to high-impact regions or features in the image, which is crucial for reconstructing high-quality textures and edges. 

\subsubsection{Loss Function}
In the Real Embedding Extractor (REE), we use MSE loss for training. It can be defined as
\begin{equation} \label{eq:mse}
  \begin{split}
  \mathcal{L}_{\text{MSE}} = \left \| f_{x}^{rep} - f_{y}^{rep} \right \|_2^2,
  \end{split}
\end{equation}
where $f_{x}^{\text{rep}}$ represents the representation embedding of the GT branch, and $f_{y}^{\text{rep}}$ represents the representation embedding of the LR branch. To optimize the overall framework, the total loss is defined as
\begin{equation} 
  \begin{split}
    &\mathcal{L}_{\text{pixel}} = \left \| I_{SR}- I_{HR} \right \|_1 , \\
    &\mathcal{L}_{\text{total}} = \mathcal{L}_{\text{pixel}}  + \lambda_1 \mathcal{L}_{\text{perceptual}} + \lambda_2 \mathcal{L}_{\text{adversarial}}.
  \end{split}
\end{equation}
where $\lambda_1$ and $\lambda_2$ denote the balancing parameters. \(\mathcal{L}_{\text{pixel}}\) is calculated using the 1-norm distance. $I_{SR}$ and $I_{HR}$ represent the super-resolution image and the high-resolution image, respectively. The perceptual loss \(\mathcal{L}_{\text{perceptual}}\) and adversarial loss \(\mathcal{L}_{\text{adversarial}}\) are defined as in~\cite{dasr}.

\section{Experiments}

\subsection{Datasets and Metrics}
We employ DIV2K~\cite{div2k}, Flickr2K~\cite{div2k}, and OutdoorSceneTraining~\cite{OST} datasets to train the DACESR network.
The LR-HR pairs are generated by the high-order degradation model~\cite{dasr}. During testing, we selected a validation set containing various types and degrees of degradation. ``Bicubic" refers to the DIV2K validation set with bicubic degradation downsampling in MATLAB. ``Level-I", ``Level-II" and ``Level-III" denote datasets with mild, medium, and severe degradations, respectively, generated using the degradation model from~\cite{dasr}. RealSR-cano~\cite{real-SR-cano} and RealSR-Nikon~\cite{real-SR-cano} consist of real-world data pairs captured by specific cameras. AIM2019-val~\cite{AIM2019-val}, provided by the AIM2019 real-world image SR challenge, is a synthetic dataset that employs realistic and unknown degradations. 
In addition, we also utilize the real-world internet dataset RealWorld38~\cite{bsrgan,realesrgan} to observe visual effects.
For measurement, we use PSNR and LPIPS~\cite{lpips} as metrics, with PSNR reflecting the fidelity of the reconstruction and LPIPS reflecting the perceptual quality of the results. The PSNR is calculated on the Y channel of YCbCr space.
\subsection{Implementation Details}

\noindent\textbf{Real Embedding Extractor:}
We randomly select 30 images from the DIV2K-Val dataset and generate 1000 distinct degradation types from the degradation space introduced in~\cite{realesrgan}. Degradations with a text similarity score greater than 0.710 are categorized as mild (top 1/4), while those below 0.297 are considered severe (bottom 1/4). For training, we use the COCO dataset~\cite{coco}, resizing each image to $512\times512$ resolution, and applying severe degradations randomly to every training image. During training, we employ the LoRA method (low-rank dimension r = 8)~\cite{lora} to fine-tune the entire REE module from DAPE~\cite{seesr} for 300K iterations, with a batch size of 8 and a learning rate of $10^{-4}$.

\noindent\textbf{The Overall Network:} Follwing~\cite{realesrgan}, we first train the MSE-based model (the teacher model of~\cite{dvmsr}) and introduce the generative adversarial training to fine-tune the GAN-based model. During training, we set the input patch size to 64 $\times$ 64 and use horizontal flipping for data augmentation. The batch size is set to 64 and the total number of iterations is 500k. The initial learning rate is set to $2\times 10^{-4}$. 
Adam optimizer with $\beta _{1} = 0.9$ and $\beta _{2} = 0.99$ is used to train the model. We balance the training loss with
$\lambda _{1} : \lambda _{2} = 1 : 0.1$. The
training process is conducted with a single NVIDIA RTX 3090 GPU.

\begin{figure*}[!htbp]
\centering
\includegraphics[width=\textwidth]{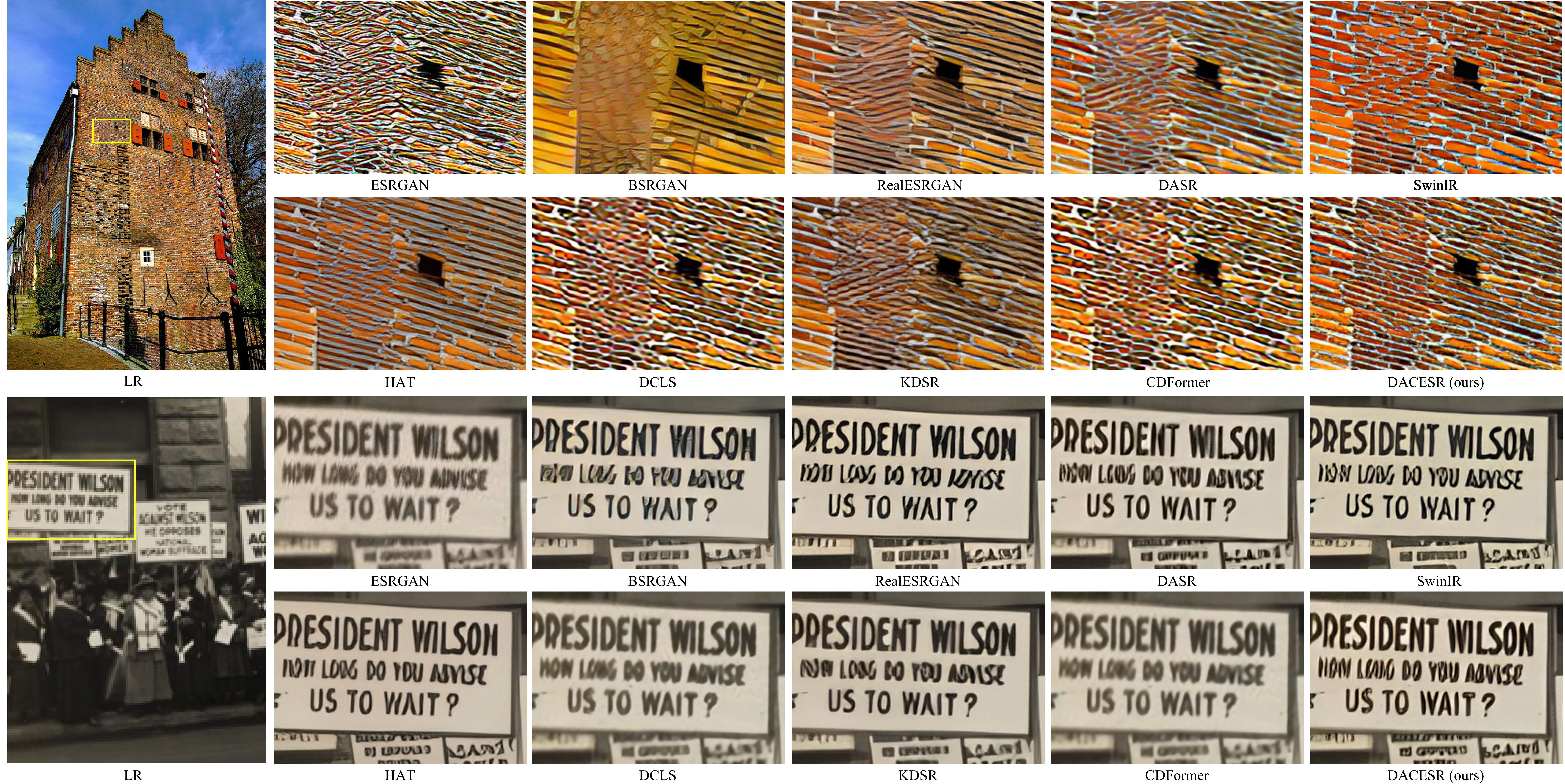}
\caption{Real-world image super-resolution results on SR $\times$ 4.}
\label{fig:visual}
\end{figure*}

\begin{table*}[!htbp]
    \centering
    \caption{Quantitative comparison of different backbone networks for real-world image super-resolution, all retrained on the same degradation space. FLOPS are measured with a $256 \times 256$ input, and all data are obtained using an NVIDIA V100 GPU. ``Mamba-based" is a Mamba-based method, derived from~\cite{dvmsr}. The best results are highlighted in bold. }
    \resizebox{\linewidth}{!}{
    \begin{tabular}{lccccccccc}
    \hline
    \multirow{2}{*}{Network}  & \multirow{2}{*}{Params(M)} & \multirow{2}{*}{FLOPS(G)} & \multirow{2}{*}{Time(ms)} & Bicubic & Level-I & Level-II & Level-III  \\
    & \multicolumn{1}{l}{}  & & & $\mathrm{PSNR}~(\uparrow)/\mathrm{LPIPS}~(\downarrow)$ & $\mathrm{PSNR}~(\uparrow)/\mathrm{LPIPS}~(\downarrow)$ & $\mathrm{PSNR}~(\uparrow)/\mathrm{LPIPS}~(\downarrow)$ & $\mathrm{PSNR}~(\uparrow)/\mathrm{LPIPS}~(\downarrow)$  \\
    \hline
    \hline
    SRResNet  & 1.52 & 166 & 113 & 28.05/0.1747 & 27.60/0.1772 & 27.34/0.2228 & 23.71/0.4419  \\
    EDSR & 1.52 & 130 & 105 & 28.26/0.1807 & 27.79/0.1834 & 27.53/0.2284 & 23.87/0.4351  \\
    SwinIR  & 11.72 & 539 & 1719 & 28.28/0.1488 & 27.78/0.1531 & 27.45/0.1854 & 23.60/0.3869  \\
    RRDB & 16.70 & 1176 & 460 & 27.92/0.1473 & 27.84/0.1569 & 27.29/0.1886 & 23.54/0.3847  \\
    DASR  & 8.07 & 184 & 142 & \textbf{28.55}/0.1696 & 27.84/0.1707 & 27.58/0.2126 & {23.93}/0.4144  \\

    KDSR  &  18.85  &  867  &  217  & 28.45/\textbf{0.1400} & 27.83/0.1493 & 27.49/0.1773 & \textbf{24.23}/\textbf{0.3420} \\
   
    Mamba-based  & 3.65 & 266 & 232 & {28.35}/0.1413& \textbf{27.94}/\textbf{0.1468}  & \textbf{27.62}/\textbf{0.1748} &  23.80/0.3538 \\
    \hline
    \end{tabular}}
    \label{tab:dasr}
\end{table*}
\subsection{Experimental Results}

\subsubsection{Quantitative Results}
Table~\ref{tab:real} presents a quantitative comparison of our method with state-of-the-art approaches~\cite{ESRGAN,bsrgan,realesrgan,swinir,dasr,HAT,kdsr,dcls,cdformer} on real-world image super-resolution. All methods use the official released weights. Except for ESRGAN, which uses bicubic degradation, the other models are trained based on real-world degradation models. The quantitative results in the table demonstrate that our DACESR consistently achieves strong performance, particularly in terms of PSNR on the Level-I and Level-II datasets, as well as lower LPIPS scores in several cases, indicating its ability to generate high-quality, perceptually pleasing images. Notably, DACESR excels on the RealSR and AIM2019-val datasets in terms of LPIPS, surpassing other methods in perceptual quality. 
Moreover, due to the outstanding performance of diffusion-based methods in super-resolution, we have compared several state-of-the-art diffusion-based approaches~\cite{stablesr, pasd, diffbir, seesr}, with quantitative results presented in Table~\ref{tab:diff}. 
The results demonstrate that our method still maintains leading performance in terms of both PSNR and LPIPS.
Overall, DACESR exhibits competitive or superior results compared to other state-of-the-art methods, especially in balancing both fidelity and perceptual quality across various degradation levels and datasets. 

\begin{table}[!htbp]
    \centering
    \caption{Quantitative comparison with diffusion-based methods. The data for the methods we compared is sourced from~\cite{seesr}. The best results are highlighted in bold.}
    \resizebox{0.8\linewidth}{!}{
    \begin{tabular}{lcccc}
    \hline
    \multirow{2}{*}{Method}  & DrealSR~\cite{drealsr} & RealSR~\cite{real-SR-cano}  \\
    &  $\mathrm{PSNR}~(\uparrow)/\mathrm{LPIPS}~(\downarrow)$ & $\mathrm{PSNR}~(\uparrow)/\mathrm{LPIPS}~(\downarrow)$  \\
    \hline
    \hline
    StableSR~\cite{stablesr}  &    28.13/0.3315 &  24.70/0.3018  \\

    PASD~\cite{pasd} &   27.00/0.3931 & 24.29/0.3435   \\

    DiffBIR~\cite{diffbir} & 26.76/0.4599 & 24.77/0.3658   \\
    SeeSR~\cite{seesr} & 28.17/0.3189 & 25.18/0.3009   \\
    
    Ours & \textbf{29.74}/\textbf{0.2803} & \textbf{27.31}/\textbf{0.2911} \\
    \hline
    \end{tabular}}
    \label{tab:diff}
\end{table}

\subsubsection{Qualitative Results}
We present the visual results of different methods on real-world low-resolution images (RealWorld38) in Fig.~\ref{fig:visual}. 
In the set of images on the left, DACESR generates more natural wall textures and colors. In the set on the right, DACESR produces clearer text, allowing us to easily recognize ``PRESIDENT WILSON" and ``US TO WAIT?". These results demonstrate that our method can produce visually pleasing outcomes with fewer artifacts and sharper edges, highlighting the great potential of DACESR in real-world scenarios.

\subsection{Ablation Study}
\subsubsection{Comparison of Different Backbones in Real-SR}
Table~\ref{tab:dasr} shows the results of different network architectures trained on the same degradation space. 
SRResNet~\cite{EDSR}, EDSR~\cite{EDSR}, RRDB~\cite{rrdb}, DASR~\cite{dasr}, and KDSR~\cite{kdsr} are CNN-based methods, while SwinIR~\cite{swinir} is a transformer-based approach. 
Although SRResNet and EDSR are more lightweight, their performance on the test set still requires improvement, particularly in terms of perceptual quality as measured by LPIPS. RRDB and DASR have relatively higher FLOPS and parameter counts, which lead to improved PSNR and LPIPS scores. This indicates a trade-off between model complexity and image quality. SwinIR achieves better performance on the test set, especially in perceptual quality (LPIPS), but it demands more inference time and higher computational complexity. 
Mamba-based achieves strong performance with minimal parameters and relatively low FLOPS. It demonstrates competitive PSNR and LPIPS values, particularly in challenging scenarios (e.g., Level-II and Level-III degradations), highlighting its ability to balance computational efficiency and perceptual quality effectively.

\begin{table*}[!htbp]
    \centering
    \caption{Quantitative Comparison of Different Conditional Information Prompts. ``Mamba-based" refers to the backbone composed of RSSB~\cite{dvmsr}. ``RE" represents the representation embedding of the low-resolution image, and ``LE" represents the logits embedding of the low-resolution image.}
    \resizebox{\linewidth}{!}{
    \begin{tabular}{lccccccc}
    \hline
    \multirow{2}{*}{Backbone} & \multirow{2}{*}{Condition} &  Bicubic & Level-I & Level-II & Level-III  \\
    &  &$\mathrm{PSNR}~(\uparrow)/\mathrm{LPIPS}~(\downarrow)$ & $\mathrm{PSNR}~(\uparrow)/\mathrm{LPIPS}~(\downarrow)$ & $\mathrm{PSNR}~(\uparrow)/\mathrm{LPIPS}~(\downarrow)$ & $\mathrm{PSNR}~(\uparrow)/\mathrm{LPIPS}~(\downarrow)$  \\
    \hline
    \hline
    Mamba-based  & None & \textbf{28.05}/0.1811 & \textbf{27.57}/0.1840 & \textbf{27.39}/0.2209 & \textbf{23.80}/\textbf{0.3938}  \\

    Mamba-based  & RE & {27.79}/\textbf{0.1546}& {27.36}/\textbf{0.1676}  & {27.08}/\textbf{0.1989} & {23.73}/{0.3975}  \\
    Mamba-based  & LE & 27.56/0.1945 & 27.23/0.1903  & 27.07/0.2227 & 23.67/0.4078   \\

    \hline
    \end{tabular}}
    \label{tab:ablation1}
\end{table*}

\begin{figure}[!htbp]
\centering
\includegraphics[width=0.45\textwidth]{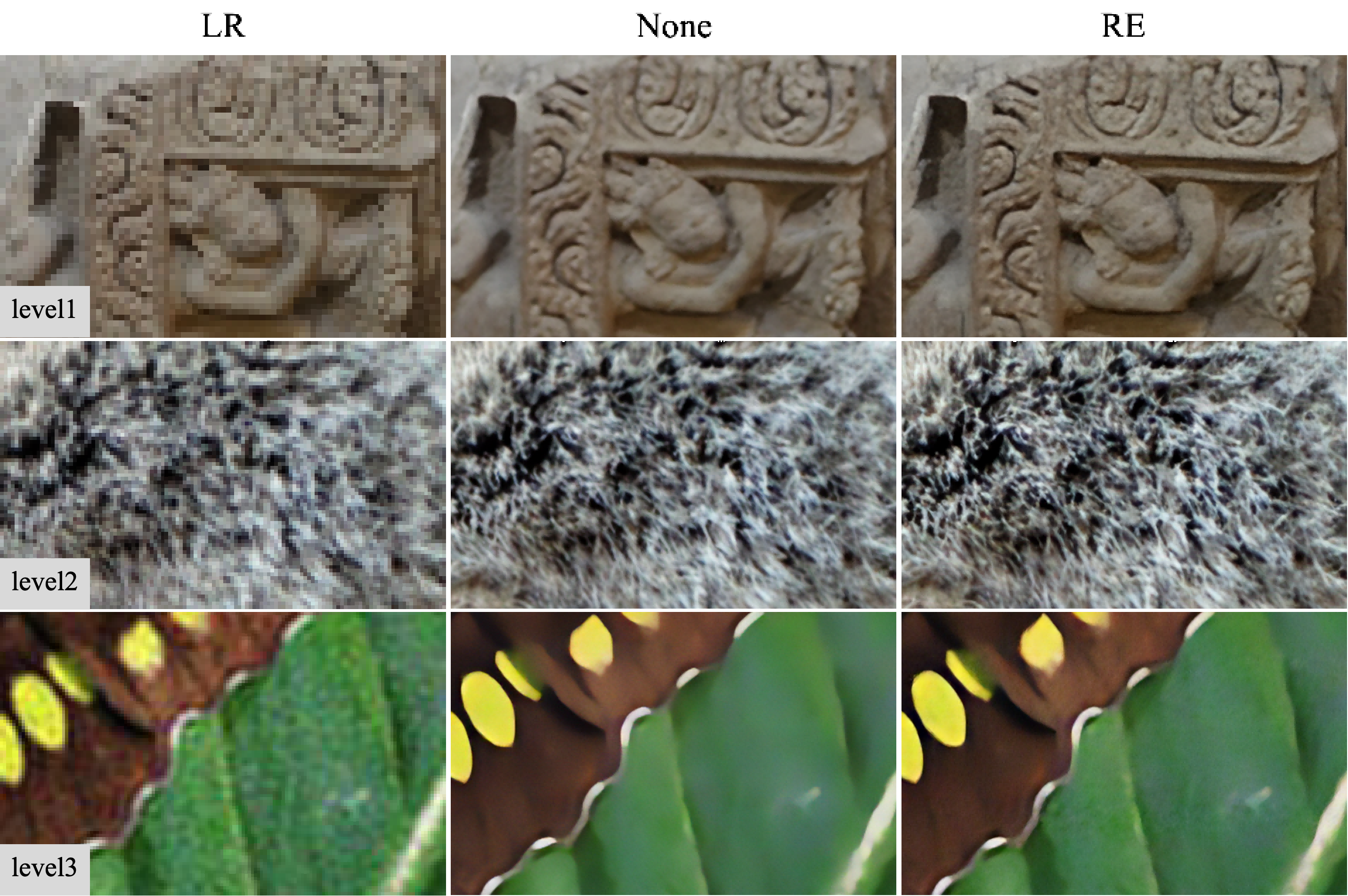}
\caption{Visual comparison of results with and without conditional information under different degradations using the same backbone network.}
\label{fig:ab_re_none}
\end{figure}
\begin{table*}[!htbp]
    \centering
    \caption{Quantitative Comparison of Different Conditional Networks. ``Mamba-based" refers to the RSSB~\cite{dvmsr}.}
    \resizebox{\linewidth}{!}{
    \begin{tabular}{lccccccc}
    \hline
    \multirow{2}{*}{Backbone} &{Condition}&  Bicubic & Level-I & Level-II & Level-III  \\
    & Network&$\mathrm{PSNR}~(\uparrow)/\mathrm{LPIPS}~(\downarrow)$ & $\mathrm{PSNR}~(\uparrow)/\mathrm{LPIPS}~(\downarrow)$ & $\mathrm{PSNR}~(\uparrow)/\mathrm{LPIPS}~(\downarrow)$ & $\mathrm{PSNR}~(\uparrow)/\mathrm{LPIPS}~(\downarrow)$  \\
    \hline
    \hline
    Mamba-based  & RAM &  27.72/0.1761 & 27.27/0.1770  & 27.08/0.2067  &  23.69/\textbf{0.3860} \\
    Mamba-based  & DAPE&  27.45/0.1720 & 27.25/0.1740  & 26.96/0.2062 & 23.68/0.3931  \\

    Mamba-based  & REE & \textbf{27.79}/\textbf{0.1546}& \textbf{27.36}/\textbf{0.1676}  & \textbf{27.08}/\textbf{0.1989} & \textbf{23.73}/{0.3975}  \\
    \hline
    \end{tabular}}
    \label{tab:ablation2}
\end{table*}

\begin{figure}[!htbp]
\centering
\includegraphics[width=0.5\textwidth]{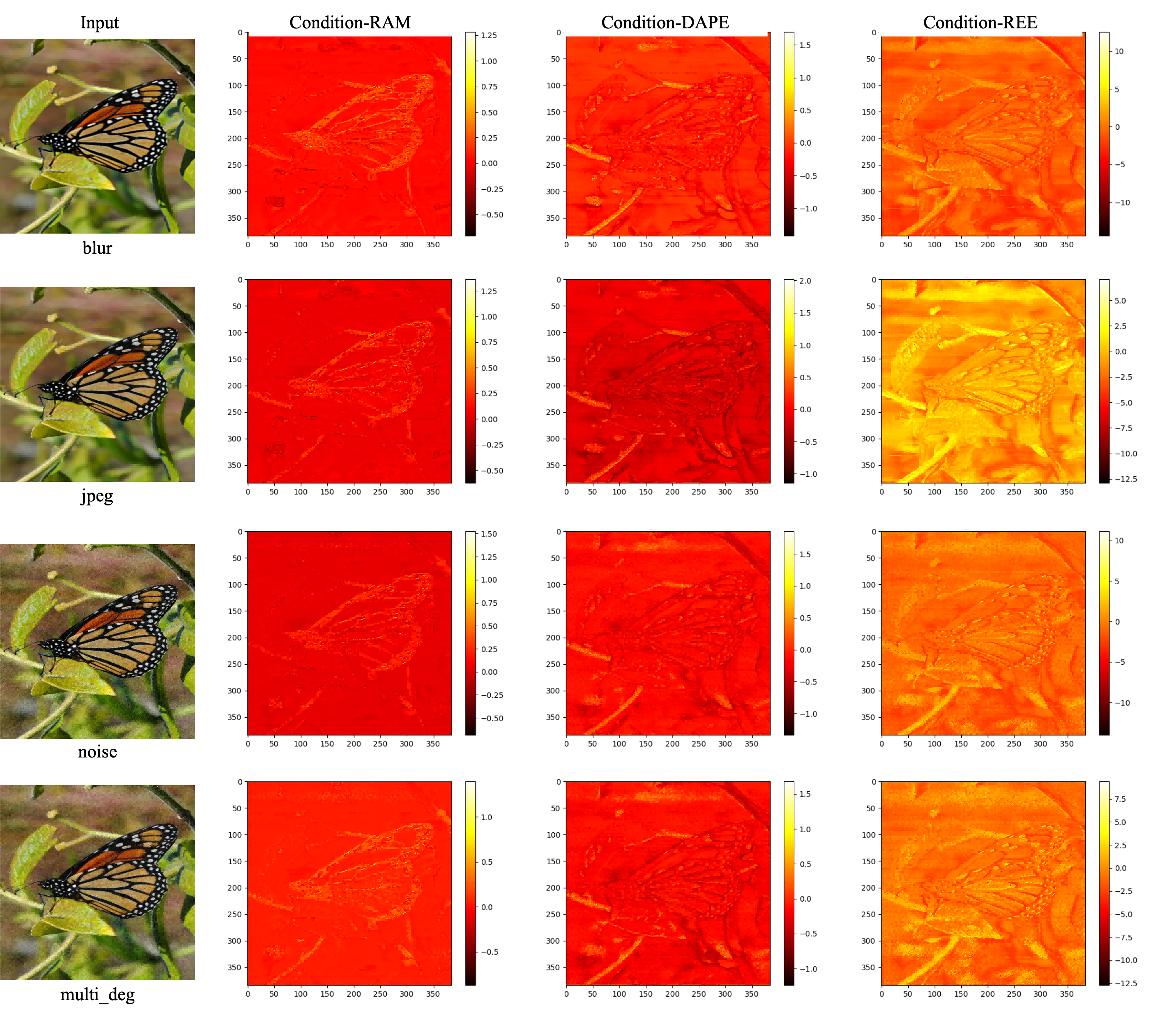}
\caption{Using images with different degradations as input, the average activation maps of the intermediate layers of each network.}
\label{fig:ab_feature_map}
\end{figure}

\subsubsection{Effectiveness of Conditional Information}
We compared the impact of different conditioning information on the super-resolution network. As shown in Table~\ref{tab:ablation1}, ``RE" represents conditioning information at the image level, involving pixel-based content, while ``LE" represents semantic-level conditioning information. The findings in~\cite{seesr} indicate that using both can ensure fidelity and visual quality of the generated images. However, in Mamba-based models, there is a significant difference. In our experiments, adding the logits embedding actually resulted in an increase in LPIPS and a decrease in PSNR. 
When using ``RE" as the conditioning information, LPIPS achieves the best performance under almost all degradation types. In contrast, when the network operates without any conditioning, PSNR reaches the highest values across all degradations. This highlights the inherent trade-off between PSNR and LPIPS metrics. As shown in Fig.~\ref{fig:ab_re_none}, it can be observed that across three different degradation types, using ``RE" as the conditioning information enables the super-resolution network to reconstruct more details compared to the ``None", resulting in outputs that are more consistent with human perceptual clarity. Therefore, LPIPS plays a critical role in evaluating real-world image super-resolution methods, demonstrating the importance of the information provided by ``RE" for achieving higher visual quality.

\subsubsection{Impact of the Conditional Network on SR Results}
Through a comparison of text similarity and visual quality, we observed that REE appears to provide descriptions of degraded images that are closer to the clean image. To study its effect on the super-resolution network, we compared experimental results by adjusting different conditioning networks. As shown in Table~\ref{tab:ablation2}, all experiments use the representation embedding of the low-resolution image as conditioning information.
When employed as a condition, DAPE achieves lower PSNR than RAM across all test benchmarks, which is consistent with the conclusion drawn from Fig.~\ref{fig:cls}. However, it exhibits better LPIPS under bicubic and Level-1 degradation, since its relative descriptive accuracy is higher under mild degradation.
REE generally outperforms RAM and DAPE across different degradation levels in both PSNR and LPIPS, except for the heaviest degradation (Level-III), where RAM achieves the best perceptual quality (lowest LPIPS). This suggests that REE is more robust overall, particularly in lighter and moderate degradations, while it maintains an advantage in fidelity under severe degradation. 
The average activation maps of different networks are illustrated in Fig.~\ref{fig:ab_feature_map}. 
The inputs are categorized into four types: blur, JPEG artifacts, noise, and a combination of multiple degradations, in order to investigate the impact of different image encoding networks on the performance of the super-resolution network. For the network using REE as the image encoder, the activation values on the butterfly wings are consistently and significantly higher than those in the background regions across all four degradation types. This indicates the network's ability to effectively focus on foreground features while suppressing irrelevant areas. Moreover, the structural details of the wings remain clearly distinguishable in the average activation maps, suggesting that it retains rich textural information while achieving semantic abstraction. The overall activation distribution is both concentrated and smooth, demonstrating that the feature extraction is not only stable but also selective, thereby facilitating the reconstruction of finer texture details.

\begin{table}[!htbp]
\centering
    \caption{Quantitative Comparison of Conditioned Information Modulator Designs.}
    \resizebox{\linewidth}{!}{
    \begin{tabular}{cccccccc}
    \hline
    \multirow{2}{*}{$\times$}  & \multirow{2}{*}{$+$}  &  Bicubic & Level-I & Level-II & Level-III  \\
    & & $\mathrm{PSNR}/\mathrm{LPIPS}$ & $\mathrm{PSNR}/\mathrm{LPIPS}$ & $\mathrm{PSNR}/\mathrm{LPIPS}$ & $\mathrm{PSNR}/\mathrm{LPIPS}$  \\
    \hline
    \hline
    \ding{52}  & \ding{56}  & \textbf{27.80}/0.1877 & 27.33/0.1858 & 27.10/0.2203 & 23.52/0.4011  \\
    
    \ding{56}  &  \ding{52} &  27.71/0.1789 & \textbf{27.38}/0.1775  & \textbf{27.12}/0.2123 & 23.58/0.3982 \\
    \ding{52}  &  \ding{52} &  27.79/\textbf{0.1546}  & 27.36/\textbf{0.1676}   & 27.08/\textbf{0.1989} & \textbf{23.73}/\textbf{0.3975}   \\

    \hline
    \end{tabular}}
    \label{tab:ablation_cfm}
\end{table}

\subsubsection{Design of Conditional Feature Modulator}
To investigate how to effectively integrate conditional information into the backbone network, we conducted several ablation studies on the Conditional Feature Modulator (CFM), as shown in Table~\ref{tab:ablation_cfm}. Compared to simply using addition or multiplication for fusion, adopting scaling and shifting operations to modulate the intermediate features of the backbone network significantly improves LPIPS performance while maintaining comparable results in PSNR.

\section{Conclusion}
We propose DACESR, which consists of a Real Embedding Extractor and a Mamba-based image super-resolution network. In this paper, first, we re-explore the capabilities of the Recognize Anything Model (RAM) on degraded images by calculating text similarity. We find that directly using contrastive learning to fine-tune RAM in the degraded space is difficult to achieve acceptable results. Second, we employ a degradation selection strategy to propose a Real Embedding Extractor (REE), which achieves significant recognition performance gain on degraded image content through contrastive learning. 
Third, we use a Conditional Feature Modulator (CFM) to incorporate the high-level information of REE for a powerful Mamba-based network that can leverage effective pixel information to restore image textures and produce visually pleasing results. Extensive experiments demonstrate the effectiveness of DACESR in producing high-quality super-resolution results in
real-world scenarios.

\section{Disscution}
Our method performs exceptionally well on Mamba-based networks, but its effectiveness on other architectures, such as CNNs, Transformers, and diffusion models, has yet to be explored. Additionally, REE, as an optimization method, serves as a conditional input to the super-resolution network, significantly enhancing image reconstruction in real-world scenarios and addressing the limitations of existing methods. However, a more powerful degradation-aware feature correction model should be proposed, and we continue to explore more flexible and efficient approaches.

\bibliographystyle{IEEEtran}
\bibliography{main}

\end{document}